\documentclass[10pt,twocolumn,letterpaper]{article}

\usepackage{makecell}
\usepackage{bm}
\usepackage{multirow}
\usepackage{pgfplots}
\pgfplotsset{compat=1.18}
\usepackage{bbding}
\usepackage[pagenumbers]{cvpr} 

\newcommand{\red}[1]{{\color{red}#1}}

\usepackage{tabularx}
\usepackage{xcolor}
\definecolor{cvprblue}{rgb}{0.21,0.49,0.74}
\usepackage[pagebackref,breaklinks,colorlinks,allcolors=cvprblue]{hyperref}

\renewcommand{\red}[1]{\textcolor{red}{#1}}
\newcommand{\blue}[1]{\textcolor{blue}{#1}}

\newlength{\RealDegPatchH}
\title{
Bridging Fidelity-Reality with Controllable One-Step Diffusion \\ for Image Super-Resolution
}

\author{First Author\\
Institution1\\
Institution1 address\\
{\tt\small firstauthor@i1.org}
\and
Second Author\\
Institution2\\
First line of institution2 address\\
{\tt\small secondauthor@i2.org}
}

\author{Hao Chen \quad Junyang Chen \quad Jinshan Pan \quad Jiangxin Dong$^{\dagger}$ \quad \\
School of Computer Science and Engineering, Nanjing University of Science and Technology\\
{\tt \url{https://github.com/Chanson94/CODSR}}
}

\begin{document}
\maketitle
\begin{abstract}
%
Recent diffusion-based one-step methods have shown remarkable progress in the field of image super-resolution, yet they remain constrained by three critical limitations: (1) inferior fidelity performance caused by the information loss from compression encoding of low-quality (LQ) inputs; (2) insufficient region-discriminative activation of generative priors; (3) misalignment between text prompts and their corresponding semantic regions. 
To address these limitations, we propose CODSR, a controllable one-step diffusion network for image super-resolution. 
First, we propose an LQ-guided feature modulation module that leverages original uncompressed information from LQ inputs to provide high-fidelity conditioning for the diffusion process.
We then develop a region-adaptive generative prior activation method to effectively enhance perceptual richness without sacrificing local structural fidelity.
Finally, we employ a text-matching guidance strategy to fully harness the conditioning potential of text prompts.
%
Extensive experiments demonstrate that CODSR achieves superior perceptual quality and competitive fidelity compared with state-of-the-art methods with efficient one-step inference.
\end{abstract}    
\section{Introduction}
\label{sec:intro}
The goal of the single image super-resolution (SISR) task is to recover a high-quality (HQ) image from its corresponding degraded low-quality (LQ) observation. 
Previous approaches, e.g., CNN-based~\cite{SRCNN,RCAN,zhang2018residual,dai2019second,sun2022shufflemixer,shi2022criteria,sun2023spatially}, Transformer-based~\cite{chen2021pre,SwinIR,zhang2022efficient,zamir2022restormer,chen2023activating,chen2023dual}, and GAN-based ones~\cite{SRGAN,BSRGAN,Real-esrgan,liang2022details}, have markedly advanced the field of SISR by leveraging local information, capturing long-range dependencies, and incorporating adversarial learning. 
Nevertheless, they still fail to generate realistic and texture-rich details from severely degraded LQ inputs.

Recent diffusion-based SISR methods~\cite{StableSR,DiffBIR,SUPIR,FaithDiff,DiT4SR} have drawn increasing attention by harnessing the strong generative priors of pretrained text-to-image diffusion models (e.g., Stable Diffusion (SD)~\cite{StableDiffusion}), achieving remarkable perceptual quality.
However, these approaches generally depend on iterative denoising over dozens or even hundreds of steps, resulting in substantial computational overhead and limiting their practical applicability. 
To address this issue, recent efforts have focused on accelerating the diffusion process by simplifying it into a single-step procedure. 
Although current one-step methods have shown encouraging results, they still suffer from three major limitations.

\begin{figure}[t]
    \centering
    \includegraphics[width=1.0\linewidth]{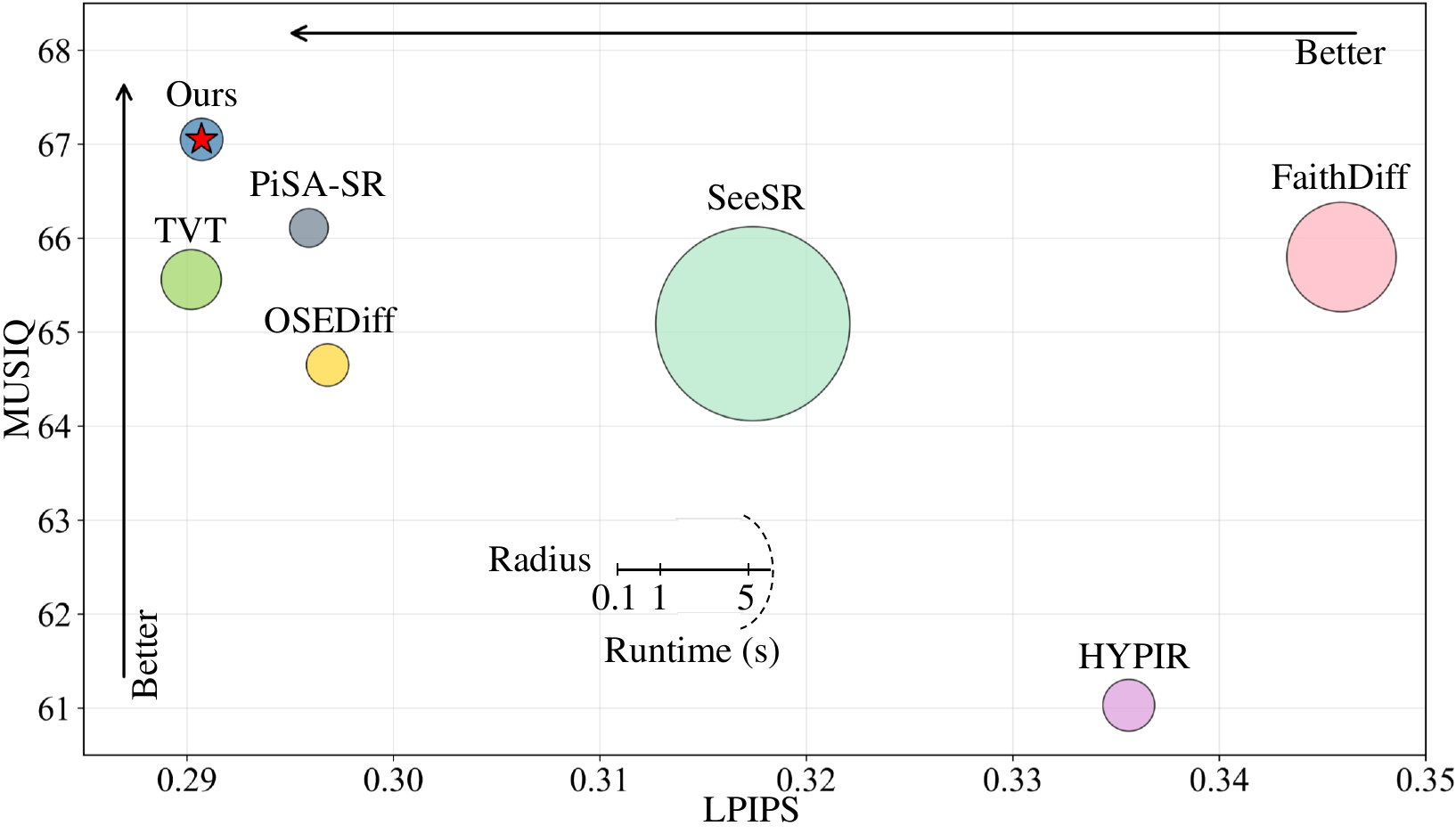}
    \vspace{-6mm} 
    \caption{
    Performance and runtime comparison among SD-based SISR methods on the DrealSR~\cite{DrealSR} benchmark. 
    CODSR achieves high-quality reconstruction at a low latency, demonstrating a better balance between fidelity and generative quality compared with existing state-of-the-art diffusion-based methods.}
    \label{fig:performance}
    \vspace{-3mm}
\end{figure}

A notable limitation of existing methods is \textbf{the inferior fidelity performance}. 
%
%
A primary factor contributing to this is the LQ information loss caused by the compression encoding. Existing methods~\cite{TVT} attempt to alleviate this by expanding the spatial resolution of features in the latent space, but the fundamental problem of information loss due to compression persists.
%
To address this, we develop an LQ-guided feature modulation module that leverages original uncompressed information from LQ inputs to modulate the diffusion process for improved fidelity performance.

In addition, 
existing one-step diffusion-based methods \cite{OSEDiff,PiSA-SR,TSD-SR,TVT} suffer from \textbf{a lack of region-discriminative activation of generative priors}.
These approaches directly feed the LQ latent into the denoising network, 
%
deviating from the pre-trained diffusion model's inherent denoising mode of recovering images from noisy latents.
This deviation limits the model's ability to extract and generate high-frequency information from noise, thereby inhibiting the full release of the model's generative potential.
Moreover, by treating all spatial regions equally, these methods overlook the varying demands for generative capacity across different image areas, often resulting in artifacts over smooth regions and insufficient detail in textured ones.
To overcome these issues, we propose a region-adaptive generative prior activation method.
%
This approach computes a Sobel-based gradient map from the grayscale LQ image to distinguish high-frequency regions from low-frequency ones and then adds adaptive noise to the high-frequency region, thereby enabling a region-aware activation of generative priors. 
The targeted approach ensures that generative priors are more effectively activated for various image areas during the reverse process, enhancing perceptual richness without compromising local structural fidelity.

Beyond the aforementioned limitations, a critical remaining challenge is \textbf{the text misalignment.} 
Text prompts offer valuable semantic guidance for SISR. 
Existing methods \cite{OSEDiff, TVT} mostly employ DAPE~\cite{SeeSR} as a text extractor. However, they overlook the issue of text misalignment, where the interaction regions of the text embeddings in the diffusion model are not spatially aligned with corresponding semantic regions of the text.
To rectify this misalignment and achieve precise semantic grounding, we propose a text-matching guidance method.
This method leverages Grounded-SAM2~\cite{GroundedSAM2} to generate text-interactive region maps, which are then used to guide and spatially align the text–image interactions throughout the diffusion process.

%

In this paper, we propose a controllable one-step diffusion network for super-resolution (CODSR), which activates the diffusion priors in a region-discriminative manner while improving fidelity by an LQ-guided feature modulation. 
Together with a text-matching guidance strategy, our CODSR shows a good capability of balancing fidelity and reality.
The contributions are summarized as follows:
\begin{itemize}

    \item We propose an LQ-guided feature modulation module that leverages original uncompressed information from LQ inputs to provide high-fidelity conditioning for the diffusion process.
    \item We present a region-adaptive generative prior activation method that effectively enhances perceptual richness without sacrificing local structural fidelity.
    \item 
    We introduce a text-matching guidance strategy that spatially aligns the text-image interactions to fully harness the conditioning potential of text prompts.    
    \item Extensive evaluations and analyses show that our CODSR achieves superior perceptual quality and competitive fidelity compared to state-of-the-art methods.
    
\end{itemize}

\begin{figure*}[ht]	
	\centering
	\includegraphics[width=\textwidth]{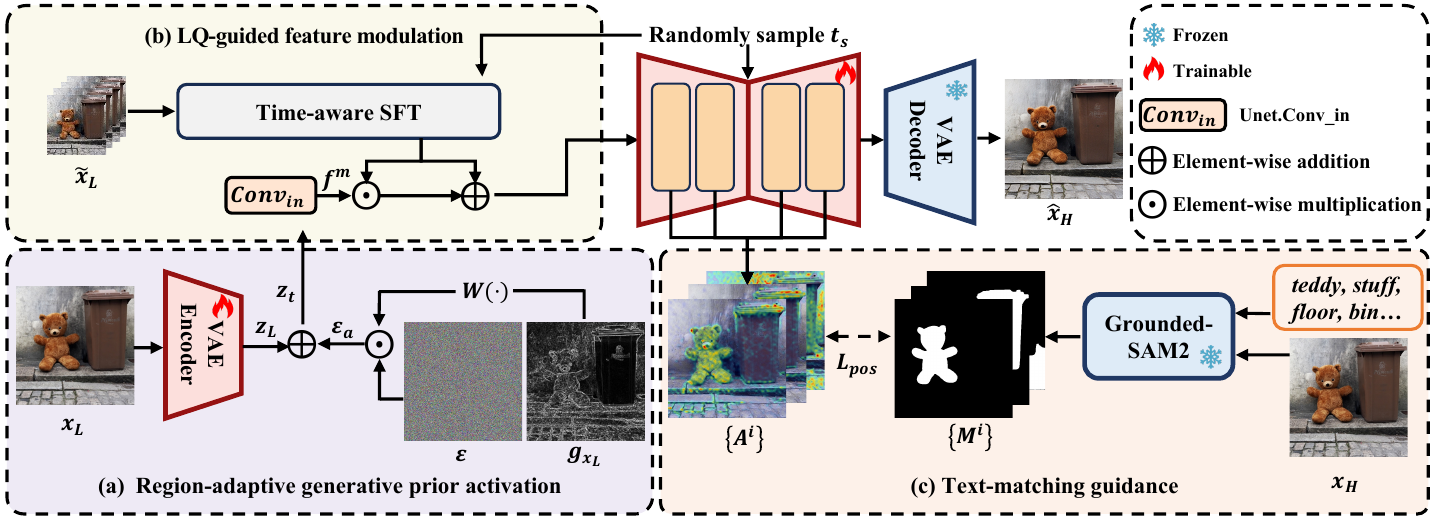}
	\caption{An overview of our CODSR. (a) The region-adaptive generative prior activation method introduces gradient-driven adaptive noise to achieve the region-aware activation of generative priors. (b) The LQ-guided feature modulation module exploits the uncompressed LQ information to modulate the diffusion process for restoring faithful structural details. (c) The text-matching guidance strategy harnesses the region maps generated by Grounded-SAM2~\cite{GroundedSAM2}, which correspond to the textual descriptions, to constrain the text–image interaction regions within the cross-attention layers, thereby enabling effective textual guidance during generation.
    }
    \label{fig:pipeline}
    \vspace{-4mm}
\end{figure*}

\section{Related Work}
\textbf{Non-generation based methods}.
Early CNN-based methods~\cite{RCAN, zhang2018residual, dai2019second, sun2022shufflemixer, shi2022criteria, sun2023spatially}, initiated by SRCNN~\cite{SRCNN} focus on exploiting local information within LQ images to reconstruct HQ images. 
However, their restricted receptive field of CNN-based models hinders the integration of global image context and impedes further quality improvements.
Transformer-based methods~\cite{chen2021pre,SwinIR,zhang2022efficient,zamir2022restormer,chen2023activating,chen2023dual} address this bottleneck by leveraging self-attention to capture long-range dependencies, leading to substantial performance improvements. 

%
%
%

\noindent
\textbf{GAN-based methods}.
Pioneering work such as SRGAN \cite{SRGAN} enhances perceptual quality by combining adversarial \cite{GAN} and perceptual losses. 
This approach is advanced by ESRGAN~\cite{ESRGAN}, which incorporates residual-in-residual dense blocks and a relativistic average discriminator to improve high-frequency detail reconstruction.
To better model real-world degradations, BSRGAN~\cite{BSRGAN} and Real-ESRGAN~\cite{Real-esrgan} subsequently employ complex degradation pipelines to synthesize more realistic training pairs, improving generalization ability on images with unknown degradations. 
Building on this progress, DASR~\cite{DASR} introduces a lightweight network to estimate degradation parameters for handling images with diverse degradation levels.
Although these GAN-based methods significantly improve perceptual quality, their adversarial training often results in instability.

%
%
%
%

\noindent
\textbf{Diffusion-based full-step methods}.
Diffusion models~\cite{dhariwal2021diffusion, DDPM, mou2024t2i} have shown remarkable potential for SISR by leveraging powerful generative priors learned from large-scale text-to-image training.
Early work~\cite{dhariwal2021diffusion, kawar2022denoising, kawar2021snips, DiffBIR} adapts pre-trained denoising diffusion probabilistic models to restore images degraded by simple operators such as bicubic downsampling.
The advent of Stable Diffusion (SD)~\cite{SD} spurs a new wave of SD-based SISR methods, often incorporating controllable adapters~\cite{mou2024t2i, zhang2023adding} for conditional guidance. 
%
%
%
%
StableSR~\cite{StableSR} introduces a time-aware encoder and feature warping to balance fidelity and perceptual quality.
DiffBIR~\cite{DiffBIR} adopts a two-stage pipeline that first removes degradation and then refines details using SD.
PASD~\cite{PASD} fuses pixel-level and semantic information through pixel-aware cross-attention. SeeSR~\cite{SeeSR} enhances semantic robustness with degradation-aware tag-style prompts. 
However, these methods rely on multi-step denoising, resulting in significant computational costs and limited practicality.

\noindent
\textbf{Diffusion-based one-step methods}.
To address this, recent studies accelerate the diffusion process by compressing it into a single-step~\cite{S3Diff}.
SinSR~\cite{SinSR} derives a deterministic sampling process from ResShift~\cite{ResShift} and adopts a consistency-preserving loss to improve the performance.
AddSR~\cite{AddSR} incorporates adversarial diffusion distillation and proposes a
prediction-based self-refinement strategy to reconstruct richer high-frequency details without introducing additional modules.
OSEDiff~\cite{OSEDiff} directly uses the LQ image as input and applies the variational score distillation~\cite{Prolificdreamer} to compress multi-step denoising into a single efficient generation step.
TSD-SR~\cite{TSD-SR} employs a target score distillation to enhance reality and a distribution-aware sampling mechanism to improve detail reconstruction.
PiSA-SR~\cite{PiSA-SR} designs dual LoRA modules to separately learn pixel-level and semantic-level objectives, effectively balancing fidelity and perceptual quality while enabling flexible control of SISR results during inference.
TVT~\cite{TVT} proposes a transfer VAE training strategy that adapts the $\times$8 VAE of SD into a $\times$4 variant through a two-stage training framework, preserving fine structural details.
Although PiSA-SR and TVT improve fidelity to some extent, the issue of information loss due to compression still persists.
HYPIR~\cite{HYPIR} leverages pretrained diffusion priors and lightweight adversarial LoRA fine-tuning, eliminating iterative sampling and distillation while achieving high-quality restoration.
However, most aforementioned methods treat different regions in an image indiscriminately, thus failing to meet the requirements for generating images with different content regions.
In contrast, we propose to explore region-discriminative activation of generative priors, in order to enhance perceptual richness without compromising local structural fidelity.

\section{Proposed Method}
The goal of this paper is to develop an effective one-step diffusion-based SR method that is capable of balancing the reality and fidelity of restored images.
We first propose a region-adaptive generative prior activation method to discriminatively add noise to regions with varying demands while preserving local fidelity.
We then introduce an LQ-guided feature modulation module that conditions the intermediate features of the U-Net on the LQ image to enhance the reconstruction fidelity.
Furthermore, we leverage Grounded-SAM2~\cite{GroundedSAM2} to identify text regions within the image, providing effective guidance for text interaction in the diffusion process and thereby mitigating text misalignment.
The overall architecture of the proposed method is shown in \Cref{fig:pipeline}. 
The following subsections elaborate on the design of each component.

\subsection{Region-adaptive generative prior activation} 
%
%
Different from existing one-step methods~\cite{OSEDiff,PiSA-SR,TSD-SR,TVT} that directly feed the LQ latent into the denoising network, we introduce the deliberate addition of Gaussian noise to the LQ latent before denoising, in order to better unleash the generative priors of diffusion models for producing visually realistic outputs.
To further achieve region-aware activation of diffusion priors, we propose a region-adaptive generative prior activation (RGPA) method that distinguishes between flat and textured regions via the Sobel gradient map and accordingly constructs spatially adaptive noise for the forward process.

 Specifically, we first encode the LQ image $\bm{x}_L$ into the latent features $\bm{z}_L$ using a VAE encoder.
Instead of directly denoising, we construct the noisy latent $\bm{z}_t$ by adding the adaptive noise $\bm{\epsilon}_a$ in a one-step forward process:
%
\begin{equation}
    \bm{z}_t = \sqrt{\bar{\alpha}_t} \bm{z}_L + \sqrt{1 - \bar{\alpha}_t} \bm{\epsilon}_a,
    \label{eq:lqForward}
\end{equation}
where 
$\bar{\alpha}_t$ is the cumulative parameter controlling the noise level at timestep $t$. 
During the reverse process, the clean latent $\bm{\hat{z}}_H$ is recovered using the noise $\bm{\epsilon}_{\theta}(\bm{z}_t,\bm{c},t)$ predicted by the U-Net, conditioned on the text embedding $\bm{c}$:
\begin{equation}
\begin{aligned}
    \bm{\hat{z}}_H &= \frac{\bm{z}_t - \sqrt{1 - \bar{\alpha}_t} \bm{\epsilon}_{\theta}(\bm{z}_t,\bm{c},t) }{\sqrt{\bar{\alpha}_t}}\\
    &= \bm{z}_L + \frac{\sqrt{1 - \bar{\alpha}_t}}{\sqrt{\bar{\alpha}_t}}( \bm{\epsilon}_a - \bm{\epsilon}_{\theta}(\bm{z}_t,\bm{c},t) )\\
    &= \bm{z}_L + w_t( \bm{\epsilon}_a - \bm{\epsilon}_{\theta}(\bm{z}_t,\bm{c},t) ),
    \label{eq:reverse}
\end{aligned}
\end{equation}
where $w_t=\frac{\sqrt{1 - \bar{\alpha}_t}}{\sqrt{\bar{\alpha}_t}}$ is a time-dependent noise coefficient.
%
%

The adaptive noise $\bm{\epsilon}_a$ is derived by a gradient-weighted process.
First, the Sobel operator is applied to compute the gradient map of $\bm{x}_L$,
denoted as $\bm{g}_{x_L}$
The noise weighting coefficients are then obtained by a mapping operator $W(\cdot)$, which performs $16 \times 16$ patch-wise averaging followed by a piecewise transformation similar to UPSR~\cite{UPSR}.
%
%
The final adaptive noise $\bm{\epsilon}_a$ is thus formulated as:
\begin{equation}
    \bm{\epsilon}_a = W(\bm{g}_{x_L}) \bm{\epsilon} ,
    \label{eq:adaptiveNoise}
\end{equation}
where $\bm{\epsilon} \sim \mathcal{N}(0, I)$. 
Additionally, to enhance the robustness of the model, a timestep $t_s \in [t_{\min}, t_{\max}]$ is randomly sampled during training, encouraging the model to generalize across various noise levels.
During inference, this design provides flexible control over the trade-off between fidelity and generative quality simply by adjusting the timestep.


\subsection{LQ-guided feature modulation}
To mitigate the LQ information loss caused by compression encoding of VAE, we propose a plug-and-play LQ-guided feature modulation (LQFM) module.
In order to preserve as much information as possible while adjusting the spatial resolution of the LQ image $\bm{x}_L$, a pixel-unshuffle operation is first applied to $\bm{x}_L$ to obtain the feature $\bm{\widetilde{x}}_L$.

To effectively utilize the information in the LQ feature $\bm{\widetilde{x}}_L$ to guide the diffusion process and achieve faithful SISR, we 
adopt a time-aware spatial feature transform (SFT) layer that fuses $\bm{\widetilde{x}}_L$ with intermediate U-Net features to bridge the representation discrepancy.
Specifically, we develop a time-aware SFT layer, conditioned on $\bm{\widetilde{x}}_L$, which performs a spatial affine transformation on the output feature $\bm{f}^m$ of the first convolutional layer in the U-Net:
\begin{equation}
    \mathrm{SFT}(\bm{f}^m \mid \bm{\widetilde{x}}_L) = (1+\lambda_t\bm{\gamma}) \odot \bm{f}^m + \lambda_t\bm{\beta},
  \label{eq: SFT_modulation}
\end{equation}
where the pair of modulation parameters $(\bm{\gamma}, \bm{\beta}) = \mathcal{M}(\bm{\widetilde{x}}_L)$ are generated by a mapping function $\mathcal{M}$, which consists of two MLP layers, $\odot$ denotes the element-wise multiplication, and 
$\lambda_t=\frac{1}{w_t}$ is a time-dependent coefficient that controls the strength of the modulation, thereby making LQFM time-aware and compatible with RGPA.

To preserve the integrity of the latent space, our method modulates the intermediate feature $\bm{f}_m$ of the U-Net, rather than the LQ latent code $\bm{z}_L$.
This targeted modulation mitigates the risk of information loss and maintains the stability of the original latent distribution, thereby ensuring a more robust denoising process (see Section~\ref{sec:ablation}).

\begin{table*}[ht]
\footnotesize
\setlength{\tabcolsep}{0pt}
\centering
\caption{
Quantitative comparison with state-of-the-art diffusion-based SR methods on four real-world test datasets.
$\uparrow$ indicates higher is better, $\downarrow$ indicates lower is better.
The best and the second-best results are highlighted in \textcolor{red}{red} and \textcolor{blue}{blue}, respectively.
}
\renewcommand\arraystretch{1.5}
\begin{tabularx}{\textwidth}{@{}l@{\hspace{1.0mm}}l@{\hspace{-1.0mm}}c@{\hspace{0.8mm}}c@{\hspace{0.8mm}}c@{\hspace{0.8mm}}c@{\hspace{-0.0mm}}c@{\hspace{0.8mm}}c@{\hspace{0.8mm}}c@{\hspace{0.8mm}}c@{\hspace{0.8mm}}c@{\hspace{0.8mm}}c@{\hspace{0.8mm}}c@{}}
\toprule
\multirow{2}{*}{Datasets} & \multirow{2}{*}{Metrics} &
\multicolumn{4}{c}{Diffusion-based full-step methods}
&
\multicolumn{6}{c}{Diffusion-based one-step methods} \\
\cmidrule(lr){3-6}\cmidrule(lr){7-12}
 &  &
~~~DiffBIR~\cite{DiffBIR} &
SeeSR~\cite{SeeSR} &
SUPIR~\cite{SUPIR} &
FaithDiff~\cite{FaithDiff} ~~ &
~~~SinSR~\cite{SinSR} &
OSEDiff~\cite{OSEDiff} &
PiSA-SR~\cite{PiSA-SR} &
TVT~\cite{TVT} &
HYPIR~\cite{HYPIR} &
Ours ~~\\
\midrule
\multirow{8}{*}{RealSR}
& PSNR (dB)$\uparrow$  & 24.97  & 25.34  & 25.16  & 25.27 & \red{\bf26.28} & 25.15 & 25.50 & \blue{25.81} & 22.89 & 25.37  \\
& SSIM$\uparrow$       & 0.6696 & 0.7294 & 0.6872 & 0.7068 & 0.7351 & 0.7341 & \blue{0.7418} & \red{\bf0.7596} & 0.6836 & 0.7284  \\
& LPIPS$\downarrow$    & 0.3486 & 0.2961 & 0.3653 & 0.2832 & 0.3189 & 0.2920 & \blue{0.2672} & \red{\bf0.2597} & 0.3046 & 0.2741  \\
& DISTS$\downarrow$    & 0.2374 & 0.2197 & 0.2439 & 0.2076 & 0.2347 & 0.2127 & \blue{0.2044} & 0.2061 & 0.2238 & \red{\bf0.2002} \\
& NIQE$\downarrow$     & 5.71   & \blue{5.35} & 6.96 & 5.46 & 6.31 & 5.64 & 5.51 & 5.92 & 5.50 & \red{\bf5.31} \\
& MUSIQ$\uparrow$      & 68.64  & 69.45  & 59.30  & 68.96  & 60.64 & 69.08 & \blue{70.15} & 69.89 & 66.42 & \red{\bf70.54} \\
& MANIQA$\uparrow$     & 0.6436 & 0.6453 & 0.5629 & \red{\bf0.6793} & 0.5384 & 0.6331 & 0.6552 & 0.6232 & 0.6510 & \blue{0.6727} \\
& CLIPIQA+$\uparrow$   & 0.5757 & 0.5762 & 0.5063 & \red{\bf0.5847} & 0.4401 & 0.5533 & 0.5635 & 0.5480 & 0.5363 & \blue{0.5768} \\
\midrule
\multirow{8}{*}{DrealSR}
& PSNR (dB)$\uparrow$  & 26.47  & 28.07  & 26.97  & 27.37 & \red{\bf28.42} & 27.92 & \blue{28.32} & 28.27 & 26.04 & 28.19 \\
& SSIM$\uparrow$      & 0.6329 & 0.7683 & 0.6942  & 0.7156 & 0.7524 & \blue{0.7835} & 0.7804 & \red{\bf0.7899} & 0.7232 & 0.7761 \\
& LPIPS$\downarrow$    & 0.4701 & 0.3174 & 0.4120 & 0.3459 & 0.3642 & 0.2968 & 0.2959 & \red{\bf0.2900} & 0.3356 & \blue{0.2919} \\
& DISTS$\downarrow$    & 0.2882 & 0.2315 & 0.2721 & 0.2386 & 0.2476 & \blue{0.2165} & 0.2169 & 0.2205 & 0.2333 & \red{\bf0.2108} \\
& NIQE$\downarrow$     & 8.14   & 6.41   & 9.03   & 6.18 & 6.95 & 6.49 & \blue{6.17} & 7.03 & 6.39 & \red{\bf5.97} \\
& MUSIQ$\uparrow$      & 62.53  & 65.09  & 53.01  & 65.80 & 55.44 & 64.65 & \blue{66.11} & 65.56 & 61.03 & \red{\bf67.05} \\
& MANIQA$\uparrow$     & 0.5987 & 0.6043 & 0.5121 & \red{\bf0.6386} & 0.4894 & 0.5895 & 0.6145 & 0.5775 & 0.5993 & \blue{0.6278} \\
& CLIPIQA+$\uparrow$   & 0.5556 & 0.5428 & 0.4452 & \red{\bf0.5649} & 0.4062 & 0.5181 & 0.5290 & 0.5226 & 0.4885 & \blue{0.5589} \\
\midrule
\multirow{4}{*}{RealPhoto60}
& NIQE$\downarrow$     & 4.70   & 4.02 & \red{\bf3.16} & 3.92 & - & 3.73 & 3.58 & 4.41 & 3.61 & \blue{3.42} \\
& MUSIQ$\uparrow$      & 63.53  & 71.76  & 70.19 & 72.02  & - & 70.46 & \blue{72.12} & 70.57 & 70.97 & \red{\bf72.72} \\
& MANIQA$\uparrow$      & 0.5881 & 0.6183 & 0.6189 & \red{\bf0.6600} & - & 0.5990 & 0.6250 & 0.5943 & 0.6240 & \blue{0.6255} \\
& CLIPIQA+$\uparrow$   & 0.5410 & \blue{0.5956} & 0.5508 & 0.5761 & - & 0.5725 & 0.5726 & 0.5700 & 0.5789 & \red{\bf0.6005} \\
\midrule
\multirow{4}{*}{RealDeg}
& NIQE$\downarrow$     & 4.22   & 4.48   & 3.96   & 3.90   & - & 3.71 & 3.59 & 4.41 & \blue{3.52} & \red{\bf3.37} \\
& MUSIQ$\uparrow$      & 58.22  & 60.10  & 51.50  & 61.24  & - & 64.97 & \blue{65.45} & 64.45 & 63.04 & \red{\bf66.39} \\
& MANIQA$\uparrow$     & 0.5487 & 0.5241 & 0.5248 & 0.5703 & - & 0.5847 & 0.6054 & 0.5548 & \blue{0.6073} & \red{\bf0.6189} \\
& CLIPIQA+$\uparrow$   & 0.4258 & 0.4315 & 0.3468 & 0.4327 & - & 0.4961 & \blue{0.5152} & 0.4821 & 0.4808 & \red{\bf0.5383} \\
\bottomrule
\end{tabularx}
\label{tab: quantitative comparisions}
\vspace{-3mm}
\end{table*}

\subsection{Text-matching guidance}
To strengthen text-image alignment and effectively leverage the conditioning potential of text prompts, we propose a text-matching guidance (TMG) strategy.
It employs Grounded-SAM2 \cite{GroundedSAM2}, an open-vocabulary segmentation model, to derive region maps corresponding to text prompts.
These maps provide explicit spatial guidance for modulating the interaction of text conditions with the latent features within the U-Net, ensuring that the textual conditioning is applied to semantically relevant image areas.


Specifically, we employ NLTK~\cite{NLTK} to remove adjectives from the prompt extracted by RAM~\cite{RAM}, as they are too abstract for precise spatial guidance.
The remaining nouns $\{ n^1, \ldots, n^N \}$, together with $\bm{x}_L$, are then fed into Grounded-SAM2~\cite{GroundedSAM2} to generate $\{ M^1, \ldots, M^N \}$, where $M^i$ indicates the binary region mask corresponding to the noun $n^i$.
%
To ensure the effectiveness of the guidance, we perform a validity check on $M^i$ and discard it if the number of active pixels is below a predefined threshold.
These masks explicitly define the regions for text interaction during the reverse process.
%
We then achieve the text-matching interaction between $n^i$ and $\bm{x}_L$ via each cross-attention layer in the U-Net, and produce the averaged attention map $A^i$ across all layers that correspond to the noun $n^i$ 
%
, supervised by the positive area loss from CoMat~\cite{CoMat}.

\subsection{Loss function}
We train the proposed network using a two-stage training strategy. To better constrain the network training in the first stage, we use the pixel-wise content loss function and the learned perceptual image patch similarity (LPIPS)~\cite{zhang2018unreasonable} loss function. In addition, the GAN-based loss function~\cite{S3Diff} is used to enhance the generation quality, particularly by improving realistic details and textures. 

%
%
%
In the second stage, we adopt a dual-LoRA training strategy inspired by PiSA-SR~\cite{PiSA-SR}.
To further enhance the semantic generation capability, semantic knowledge is distilled from a pretrained model using VSD loss~\cite{OSEDiff}.
The LoRA adapters trained in the first stage are frozen, while additional LoRA adapters are integrated exclusively into the cross-attention layers of the U-Net to enhance text alignment. 
This stage focuses on optimizing the following objective to address potential text misalignment issues:
\begin{equation}
    \mathcal{L} = \mathcal{L}_{\text{OSEDiff}}
    +\eta_\text{pos} \mathcal{L}_{\text{pos}},
    \label{eq:loss_stage2}
\end{equation}
where $\mathcal{L}_{\text{OSEDiff}}$ denotes the loss function used in OSEDiff~\cite{OSEDiff}, $\eta_\text{pos}$ is a weight parameter that is empirically set to $1$ in the proposed method and $\mathcal{L}_{\text{pos}}$ is the positive area loss used in CoMat~\cite{CoMat} to supervise text-matching interaction.
\begin{figure*}[t]
\centering
\captionsetup{font=small}


\begin{subfigure}[t]{0.345\textwidth} 
  \vspace{0pt}
  \centering
  \includegraphics[width=\linewidth]{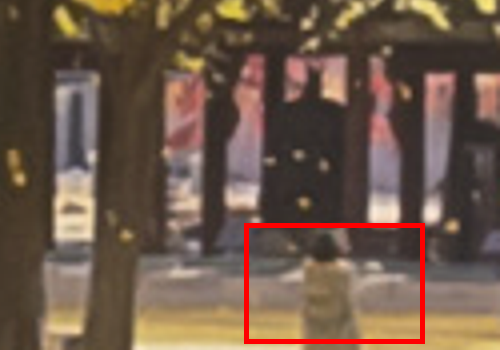}
  \vspace{-12pt}\caption*{\small LQ image from RealSR~\cite{RealSR}}
\end{subfigure}
\hfill
\begin{subfigure}[t]{0.65\textwidth} 
  \vspace{0pt}
  \centering
  \begin{subfigure}[t]{0.245\linewidth}
    \centering\includegraphics[width=\linewidth]{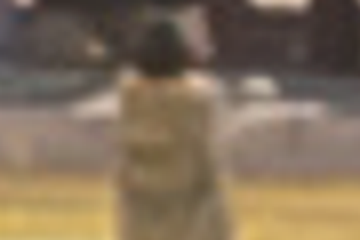}
    \vspace{-12pt}\caption*{\small(a1) LQ patch}
  \end{subfigure}\hfill
  \begin{subfigure}[t]{0.245\linewidth}
    \centering\includegraphics[width=\linewidth]{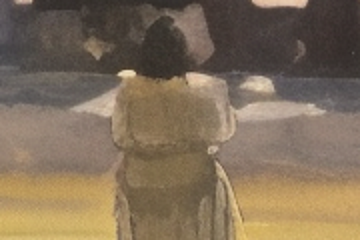}
    \vspace{-12pt}\caption*{\small(b1) DiffBIR~\cite{DiffBIR}}
  \end{subfigure}\hfill
  \begin{subfigure}[t]{0.245\linewidth}
    \centering\includegraphics[width=\linewidth]{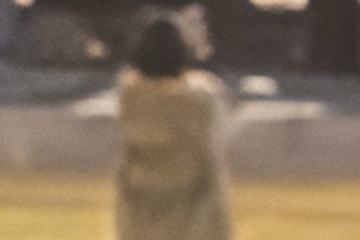}
    \vspace{-12pt}\caption*{\small(c1) SUPIR~\cite{SUPIR}}
  \end{subfigure}\hfill
  \begin{subfigure}[t]{0.245\linewidth}
    \centering\includegraphics[width=\linewidth]{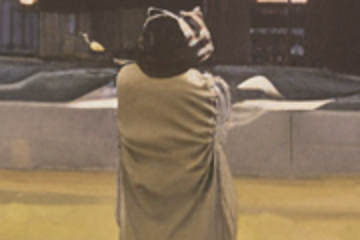}
    \vspace{-12pt}\caption*{\small(d1) FaithDiff~\cite{FaithDiff}}
  \end{subfigure}

  \vspace{0.5pt}

  \begin{subfigure}[t]{0.245\linewidth}
    \centering\includegraphics[width=\linewidth]{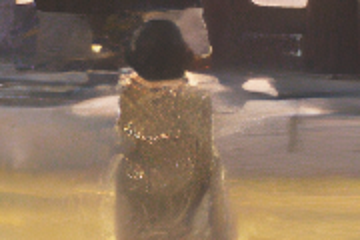}
    \vspace{-12pt}\caption*{\small(e1) SinSR~\cite{SinSR}}
  \end{subfigure}\hfill
  \begin{subfigure}[t]{0.245\linewidth}
    \centering\includegraphics[width=\linewidth]{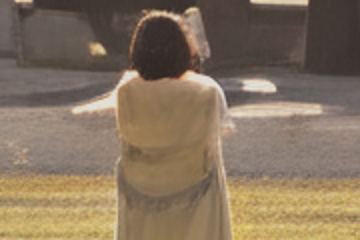}
    \vspace{-12pt}\caption*{\small(f1) OSEDiff~\cite{OSEDiff}}
  \end{subfigure}\hfill
  \begin{subfigure}[t]{0.245\linewidth}
    \centering\includegraphics[width=\linewidth]{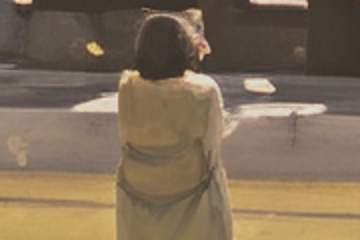}
    \vspace{-12pt}\caption*{\small(g1) PiSA-SR~\cite{PiSA-SR}}
  \end{subfigure}\hfill
  \begin{subfigure}[t]{0.245\linewidth}
    \centering\includegraphics[width=\linewidth]{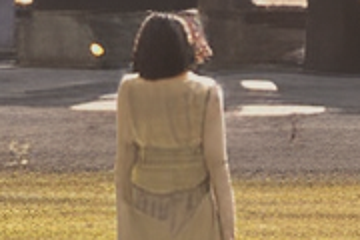}
    \vspace{-12pt}\caption*{\small(h1) Ours}
  \end{subfigure}
\end{subfigure}

\vspace{1pt} 

\begin{subfigure}[t]{0.345\textwidth} 
  \vspace{0pt}
  \centering
  \includegraphics[width=\linewidth]{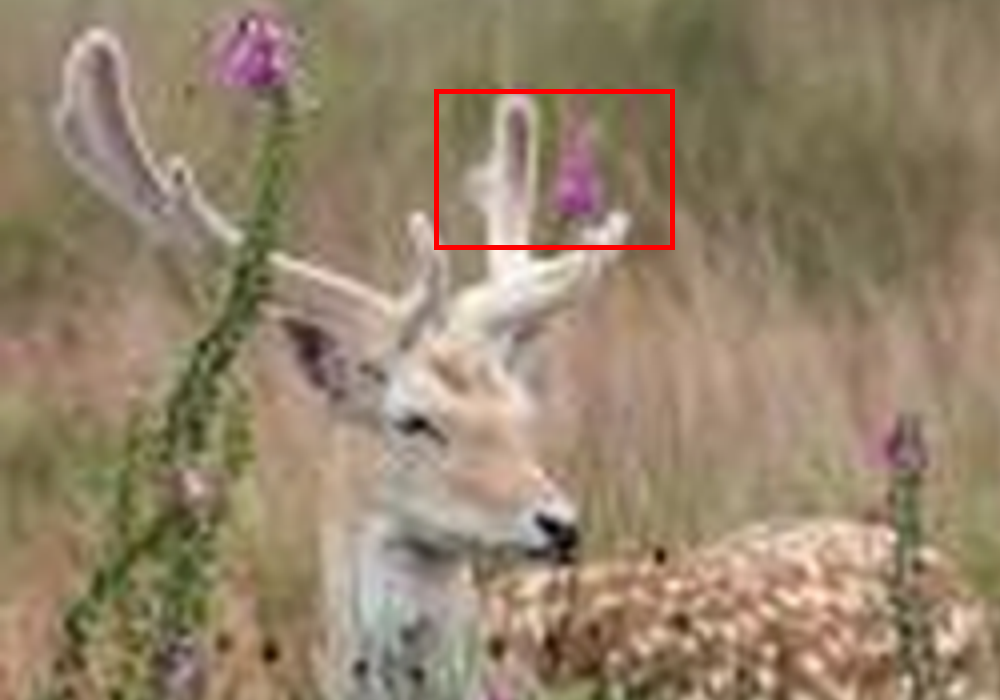}
  \vspace{-12pt}\caption*{\small LQ image from RealLQ250~\cite{ai2024dreamclear}}
\end{subfigure}
\hfill
\begin{subfigure}[t]{0.65\textwidth} 
  \vspace{0pt}
  \centering
  \begin{subfigure}[t]{0.245\linewidth}
    \centering\includegraphics[width=\linewidth]{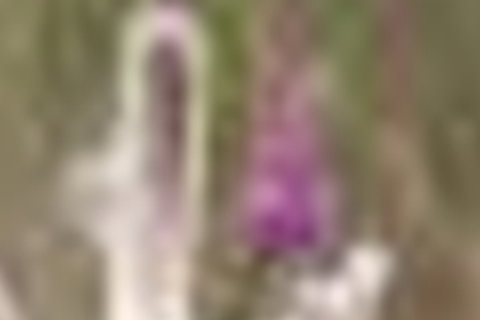}
    \vspace{-12pt}\caption*{\small(a3) LQ patch}
  \end{subfigure}\hfill
  \begin{subfigure}[t]{0.245\linewidth}
    \centering\includegraphics[width=\linewidth]{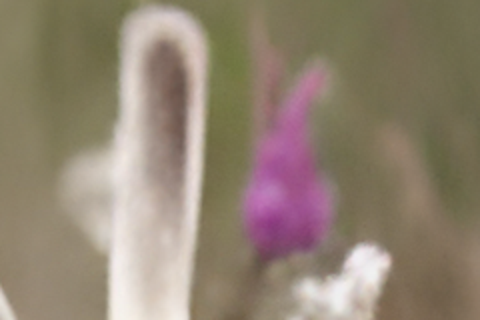}
    \vspace{-12pt}\caption*{\small(b3) \scriptsize{Real-ESRGAN~\cite{Real-esrgan}}}
  \end{subfigure}\hfill
  \begin{subfigure}[t]{0.245\linewidth}
    \centering\includegraphics[width=\linewidth]{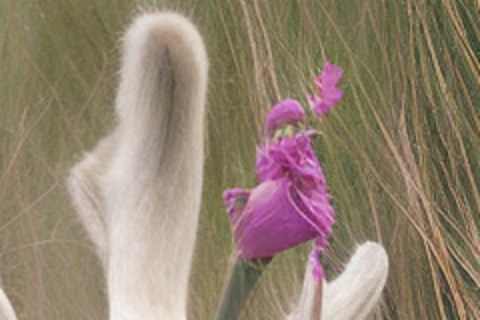}
    \vspace{-12pt}\caption*{\small(c3) DiffBIR~\cite{DiffBIR}}
  \end{subfigure}\hfill
  \begin{subfigure}[t]{0.245\linewidth}
    \centering\includegraphics[width=\linewidth]{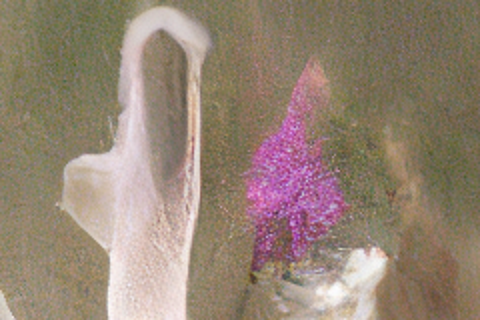}
    \vspace{-12pt}\caption*{\small(d3) SinSR~\cite{SinSR}}
  \end{subfigure}
  \vspace{0.5pt}
  \begin{subfigure}[t]{0.245\linewidth}
    \centering\includegraphics[width=\linewidth]{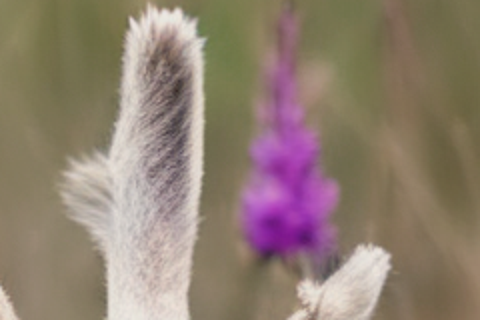}
    \vspace{-12pt}\caption*{\small(e3) OSEDiff~\cite{OSEDiff}}
  \end{subfigure}\hfill
  \begin{subfigure}[t]{0.245\linewidth}
    \centering\includegraphics[width=\linewidth]{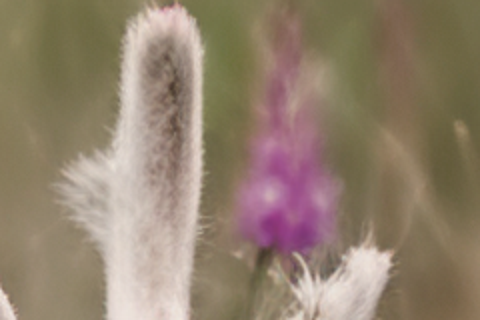}
    \vspace{-12pt}\caption*{\small(f3) PiSA-SR~\cite{PiSA-SR}}
  \end{subfigure}\hfill
  \begin{subfigure}[t]{0.245\linewidth}
    \centering\includegraphics[width=\linewidth]{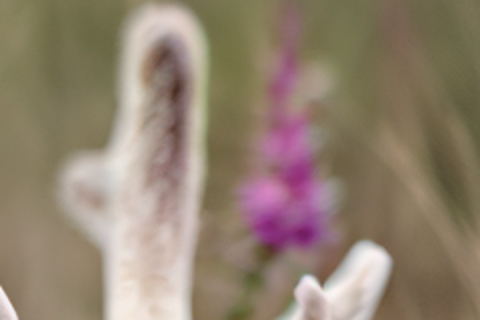}
    \vspace{-12pt}\caption*{\small(g3) HYPIR~\cite{HYPIR}}
  \end{subfigure}\hfill
  \begin{subfigure}[t]{0.245\linewidth}
    \centering\includegraphics[width=\linewidth]{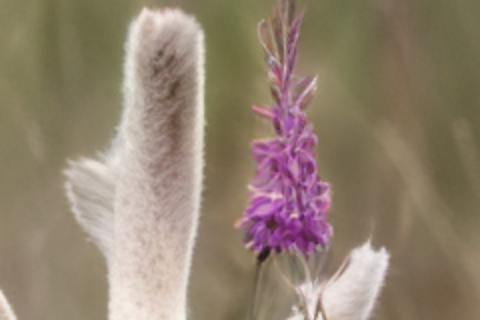}
    \vspace{-12pt}\caption*{\small(h3) Ours}
  \end{subfigure}
\end{subfigure}

\vspace{1pt} 

\begin{subfigure}[t]{0.345\textwidth} 
  \vspace{0pt}
  \centering
  \includegraphics[width=\linewidth]{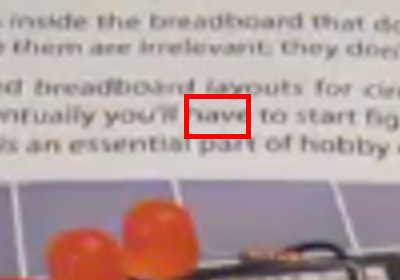}
  \vspace{-12pt}\caption*{\small LQ image from RealLR200~\cite{SeeSR}}
\end{subfigure}
\hfill
\begin{subfigure}[t]{0.65\textwidth} 
  \vspace{0pt}
  \centering
  \begin{subfigure}[t]{0.245\linewidth}
    \centering\includegraphics[width=\linewidth]{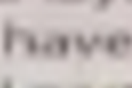}
    \vspace{-12pt}\caption*{\small(a2) LQ patch}
  \end{subfigure}\hfill
  \begin{subfigure}[t]{0.245\linewidth}
    \centering\includegraphics[width=\linewidth]{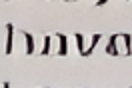}
    \vspace{-12pt}\caption*{\small(b2) \scriptsize{Real-ESRGAN~\cite{Real-esrgan}}}
  \end{subfigure}\hfill
  \begin{subfigure}[t]{0.245\linewidth}
    \centering\includegraphics[width=\linewidth]{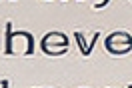}
    \vspace{-12pt}\caption*{\small(c2) DiffBIR~\cite{DiffBIR}}
  \end{subfigure}\hfill
  \begin{subfigure}[t]{0.245\linewidth}
    \centering\includegraphics[width=\linewidth]{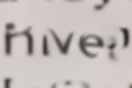}
    \vspace{-12pt}\caption*{\small(d2) SUPIR~\cite{SUPIR}}
  \end{subfigure}
  \vspace{0.5pt}
  \begin{subfigure}[t]{0.245\linewidth}
    \centering\includegraphics[width=\linewidth]{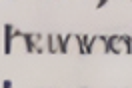}
    \vspace{-12pt}\caption*{\small(e2) OSEDiff~\cite{OSEDiff}}
  \end{subfigure}\hfill
  \begin{subfigure}[t]{0.245\linewidth}
    \centering\includegraphics[width=\linewidth]{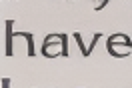}
    \vspace{-12pt}\caption*{\small(f2) PiSA-SR~\cite{PiSA-SR}}
  \end{subfigure}\hfill
  \begin{subfigure}[t]{0.245\linewidth}
    \centering\includegraphics[width=\linewidth]{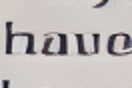}
    \vspace{-12pt}\caption*{\small(g2) TVT~\cite{TVT}}
  \end{subfigure}\hfill
  \begin{subfigure}[t]{0.245\linewidth}
    \centering\includegraphics[width=\linewidth]{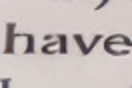}
    \vspace{-12pt}\caption*{\small(h2) Ours}
  \end{subfigure}
\end{subfigure}

\caption{Qualitative comparisons of different methods on the RealSR~\cite{RealSR} dataset, RealLQ250~\cite{ai2024dreamclear} dataset and RealLR200~\cite{SeeSR} dataset. 
%
%
Compared to competing methods, our approach generates a more realistic image with fine-scale structures and details.
Please zoom in for a better view.
}
\label{fig: qual_comp_1}
\vspace{-3mm}
\end{figure*}

\section{Experimental Results}

\subsection{Experimental settings}
\textbf{Training datasets.}
Following the protocols~\cite{SeeSR,OSEDiff,PiSA-SR}, we collect a training dataset including the images from LSDIR~\cite{li2023lsdir}, DIV2K~\cite{DIV2K}, Flicker2K \cite{Flicker2k}, DIV8K~\cite{div8k} and the first 10K images from FFHQ~\cite{karras2019style}. 
The corresponding LQ images are synthesized from their HQ counterparts through the complex degradation model in Real-ESRGAN~\cite{Real-esrgan}.

\noindent
\textbf{Testing datasets.}
We evaluate our method on four real-world datasets, including RealSR~\cite{RealSR}, DrealSR~\cite{DrealSR}, RealPhoto60~\cite{SUPIR} and RealDeg~\cite{FaithDiff}.
For RealSR and DRealSR, the LQ-HQ pairs involve a $\times 4$ super-resolution task from $128\times128$ to $512\times512$. While for RealPhoto60, the LQ images ($512\times512$) are upscaled by $\times 2$ to $1024\times1024$. 
In addition, RealDeg~\cite{FaithDiff} contains 238 images of old photographs, classic film stills, and social media photos under diverse real-world degradation types. 

\noindent
\textbf{Compared methods.}
We compare our method with representative approaches, including full-step diffusion-based methods (i.e., DiffBIR~\cite{DiffBIR}, SeeSR~\cite{SeeSR}, SUPIR~\cite{SUPIR}, and FaithDiff~\cite{FaithDiff}) 
and one-step diffusion-based methods (i.e., SinSR~\cite{SinSR}, OSEDiff~\cite{OSEDiff}, TVT~\cite{TVT}, PiSA-SR~\cite{PiSA-SR}, and HYPIR~\cite{HYPIR}).
In addition, we also include GAN-based super-resolution methods such as RealESRGAN~\cite{Real-esrgan} and BSRGAN~\cite{BSRGAN} for comprehensive comparison, as detailed in the supplemental material.
For a fair comparison, all reported results for competing methods are produced using their publicly available official implementations and pretrained models.

\noindent
\textbf{Evaluation metrics.}
We evaluate all methods
using full-reference and no-reference image quality metrics.
For full-reference assessment, we employ PSNR and SSIM \cite{wang2004image}, computed on the Y channel in YCbCr space to evaluate reconstruction fidelity, alongside LPIPS~\cite{zhang2018unreasonable} and DISTS \cite{ding2020image} to measure perceptual similarity.
%
The no-reference metrics include NIQE \cite{zhang2015feature}, MUSIQ~\cite{ke2021musiq}, MANIQA-pipal~\cite{yang2022maniqa}, and CLIPIQA+ \cite{wang2023exploring}, which estimate perceptual quality without references.

\noindent
\textbf{Implementation details.}
Our implementation is built upon the SD 2.1-base~\cite{SD}.
Similar to OSEDiff~\cite{OSEDiff}, we optimize LoRA~\cite{LoRA} adapters within the VAE encoder and the U-Net network while keeping the VAE decoder fixed. 
The LoRA ranks are configured as $4$ for the VAE encoder and $16$ for the U-Net, respectively.
We use RAM~\cite{RAM} for prompt extraction during training and DAPE~\cite{SeeSR} at inference.
Our model is trained using 4 NVIDIA 4090 GPUs with a batch size of $16$, using the AdamW optimizer~\cite{loshchilov2017decoupled} with a learning rate of $5e^{-5}$.
In the first-stage training, we employ the GAN loss used in S3Diff~\cite{S3Diff}. In the second stage, the VSD loss is incorporated with a weighting coefficient of 2.
More experimental results are included in the supplemental material.

\subsection{Comparison with the state of the art}

\textbf{Quantitative evaluations against diffusion-based full-step methods.}
We first compare our approach with diffusion-based full-step methods~\cite{DiffBIR,SeeSR,SUPIR,FaithDiff} in Table~\ref{tab: quantitative comparisions}.
Our method performs the best in terms of the all full-reference metrics (PSNR, SSIM, LPIPS, and DISTS) as well as the no-reference metrics NIQE and MUSIQ while delivering comparable results on other metrics.
Notably, our method improves the MUSIQ by at least $1.96$ and $5.15$ on the datasets of DRealSR and RealDeg, respectively.

\noindent
\textbf{Quantitative evaluations against diffusion-based one-step methods.}
We then evaluate the proposed approach against diffusion-based one-step methods~\cite{SinSR,OSEDiff,PiSA-SR,TVT,HYPIR}.
The quantitative results in Table~\ref{tab: quantitative comparisions} show that our method outperforms all competing approaches across all no-reference metrics.
The NIQE, MUSIQ, and CLIPIQA+ of our method is at least $0.15$, $0.39$, and $0.0133$ higher than the competing methods across all datasets.

\noindent
\textbf{Qualitative results.}
Figure~\ref{fig: qual_comp_1} shows the visual comparisons.
For the example from RealSR~\cite{RealSR}, the full-step methods~\cite{DiffBIR,SUPIR,FaithDiff} cannot effectively recover a clear figure viewed from the behind that is consistent with the LQ input.
The result generated by~\cite{SinSR} exhibits obvious artificial textures.
Compared to these competing methods, our approach yields a clearer result with a more realistic background and better-defined foreground.
%
%
Figure~\ref{fig: qual_comp_1} also shows an example from RealLQ250~\cite{ai2024dreamclear}, our method further demonstrates a superior capability for storing high-quality images, e.g., 
%
%
clear and natural flowers and antlers.
For the example from RealLR200~\cite{SeeSR}, the methods~\cite{Real-esrgan,SUPIR,OSEDiff} struggle to recover characters from the LQ input,
while ~\cite{DiffBIR,PiSA-SR,TVT} introduce background noise or structurally distorted characters, leading to compromised readability.
In contrast, our method faithfully restores text with sharper edges, continuous strokes, and minimal background artifacts, yielding a highly legible result.

\begin{table}[!t]
\caption{Effectiveness of each module in our proposed network. All methods are trained using the same settings as the proposed framework for fair comparison.}
\vspace{-2mm}
\centering
\resizebox{1.0\columnwidth}{!}{
\begin{tabular}{lccccccccc}
\toprule
\multicolumn{1}{c}{} &
  \multirow{2}{*}{\begin{tabular}[c]{@{}c@{}}RGPA\end{tabular}} &
  \multirow{2}{*}{\begin{tabular}[c]{@{}c@{}}LQFM\end{tabular}} &
  \multirow{2}{*}{\begin{tabular}[c]{@{}c@{}}TMG\end{tabular}} &
  \multicolumn{3}{c}{DrealSR~\cite{DrealSR}} \\ \cline{5-7}
\multicolumn{1}{c}{} &                             &                             &                             & LPIPS$\downarrow$  & MUSIQ$\uparrow$ & CLIPIQA+$\uparrow$ \\ \hline
$\text{Base}$                 & \XSolidBrush & \XSolidBrush   & \XSolidBrush   & 0.2906 & 65.89 & 0.5038              \\
$\text{Base}_\text{w/ RGPA}$                & \Checkmark   & \XSolidBrush & \XSolidBrush   & 0.2914  & 66.26 & 0.5109             \\
$\text{Base}_\text{w/ LQFM}$                  & \XSolidBrush   & \Checkmark   & \XSolidBrush  & \textbf{0.2876}   & 65.90  & 0.5106             \\
$\text{Base}_\text{w/ RGPA\&LQFM}$                  & \Checkmark   & \Checkmark   & \XSolidBrush & 0.2902  & 66.27  & 0.5213             \\
$\text{Ours}$                 & \Checkmark   & \Checkmark   & \Checkmark   & 0.2919      & \textbf{67.05}   & \textbf{0.5589}            \\ \bottomrule
\end{tabular}}
\vspace{-2mm}
\label{tab:ablation_total}
\end{table}

\section{Analysis and Discussion}
\label{sec:ablation}


\begin{figure}[t]
    \scriptsize
    \centering
    \setlength{\tabcolsep}{1pt} 
    \renewcommand{\arraystretch}{0.9} 
    \begin{tabular}{cccc}
        \includegraphics[width=0.24\columnwidth,height=0.3\columnwidth]{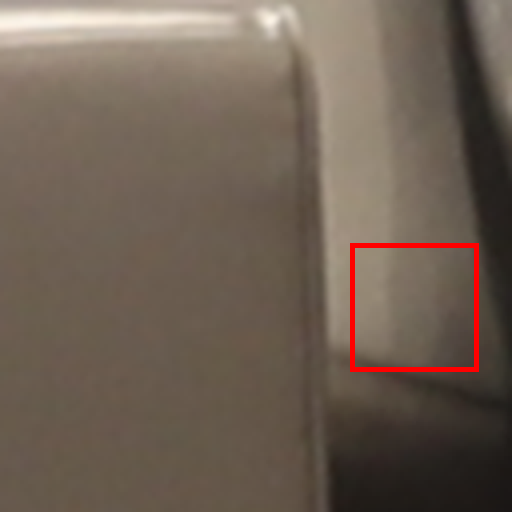} &
        \includegraphics[width=0.24\columnwidth,height=0.3\columnwidth]{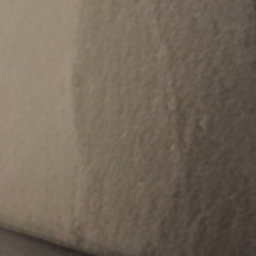} &
        \includegraphics[width=0.24\columnwidth,height=0.3\columnwidth]{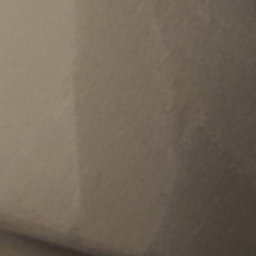} &
        \includegraphics[width=0.24\columnwidth,height=0.3\columnwidth]{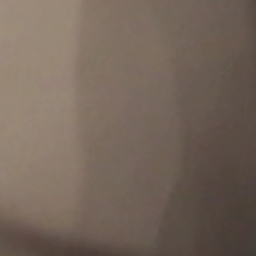} \\
        (a) $\text{LQ image}$ & (b) \emph{Base$_{\text{w/}~\bm{z}_L + \bm{\epsilon}}$} & (c) \emph{Base$_{\text{w/ RGPA}}$} & (d) GT \\
    \end{tabular}
    \vspace{-2mm}
    \caption{
    Comparisons between \emph{Base$_{\text{w/}~\bm{z}_L + \bm{\epsilon}}$} and \emph{Base$_{\text{w/ RGPA}}$} on the DrealSR~\cite{DrealSR} benchmark. While \emph{Base$_{\text{w/}~\bm{z}_L + \bm{\epsilon}}$} introduces unpleasant details in smooth areas, our method achieves more faithful reconstructions.}
    \label{fig:ablation_ganm}
    \vspace{-3mm}
\end{figure}

\begin{table}[h]
  \centering
  \caption{Quantitvative comparison of different noise strategies on the DrealSR~\cite{DrealSR} benchmark.}
  \label{tab:ablation_ganm}
  \vspace{-2mm}
  \resizebox{1.0\columnwidth}{!}{
  \begin{tabular}{@{}l c c c c@{}}
    \toprule
    & PSNR $\uparrow$ & LPIPS $\downarrow$ & MUSIQ $\uparrow$ & CLIPIQA+ $\uparrow$ \\
    \midrule
    $\text{Base}$ & \textbf{28.39} & \textbf{0.2906} & 65.89 & 0.5038 \\
    $\text{Base}_{\text{w/}~\bm{x}_L + \bm{\epsilon}_a}$ & 28.29 & 0.2917 & 65.84 & 0.5020 \\
    $\text{Base}_{\text{w/}~\bm{z}_L + \bm{\epsilon}}$ & 28.04 &  0.2974 & \textbf{66.48} & \textbf{0.5133} \\
    $\text{Base}_\text{w/ RGPA}$ & 28.24 & 0.2914  & 66.26 & 0.5109 \\
    \bottomrule
  \end{tabular}}
  \vspace{-2mm}
\end{table}

\noindent
\textbf{Effect of region-adaptive generative prior activation.} 
To validate the effectiveness of the proposed RGPA, we first compare a baseline method stripped of all the proposed modules of RGPA, LQFM, and TMG (\emph{Base} for short) against a variant incorporating only the RGPA (\emph{Base$_{\text{w/ RGPA}}$} for short).
Table~\ref{tab:ablation_total} shows the quantitative comparison, where we find that employing RGPA facilitates the release of generative priors for enhanced perceptual quality, cf. 65.89/0.5038 for \emph{Base} vs. 66.27/0.5109 for \emph{Base$_{\text{w/ RGPA}}$} in terms of MUSIQ/CLIPIQA+.

To further analyze the effect of RGPA, we compare \emph{Base$_{\text{w/ RGPA}}$} with two baselines that individually add an adaptive noise $\bm{\epsilon}_a$ to the input LQ image $\bm{x}_L$ (\emph{Base$_{\text{w/}~\bm{x}_L + \bm{\epsilon}_a}$} for short) or replace the adaptive noise with standard Gaussian noise $\bm{\epsilon}$ in the latent space (\emph{Base$_{\text{w/}~\bm{z}_L + \bm{\epsilon}}$} for short).
As shown in Table~\ref{tab:ablation_ganm}, adding noise $\bm{\epsilon}_a$ to $\bm{x}_L$ fails to enhance the generative capability, since perturbing the image space does not align the noise pathway between the forward and reverse processes in the latent space.
Although adding standard Gaussian noise $\bm{\epsilon}$ to the latent code $\bm{z}_L$ can enhance generative prior utilization, it significantly compromises fidelity, introducing artifacts in flat regions, as shown in \Cref{fig:ablation_ganm}(b).
In contrast, the method with our proposed RGPA achieves a superior balance, enabling region-adaptive activation of generative priors while faithfully preserving local structures (\Cref{fig:ablation_ganm}(c)).

\begin{figure*}[t]
  \centering
  \setlength{\tabcolsep}{1pt}        
  \renewcommand{\arraystretch}{0.92}

  \newlength{\imgw}
  \setlength{\imgw}{\dimexpr(\textwidth - 8\tabcolsep)/5\relax}
  \resizebox{\textwidth}{!}{  
  \begin{tabular}{@{}ccccc@{}} 
    \includegraphics[width=\imgw]{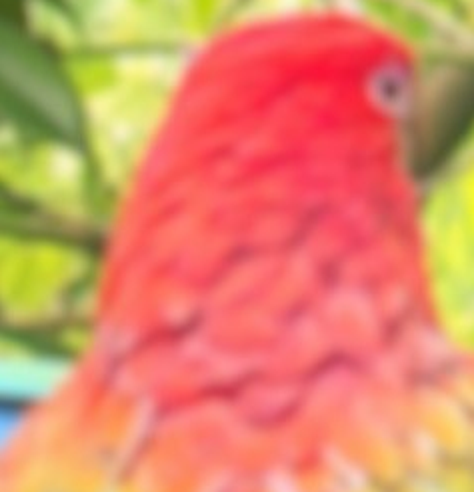} &
    \includegraphics[width=\imgw]{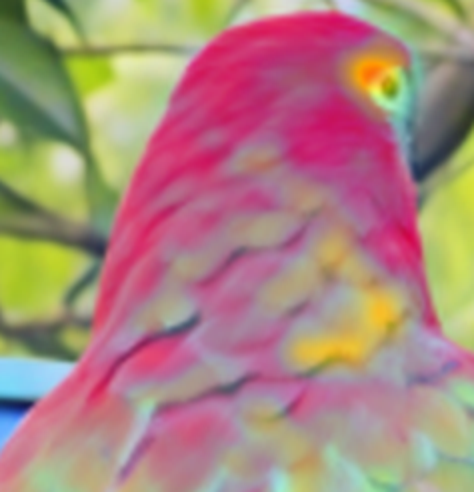} &
    \includegraphics[width=\imgw]{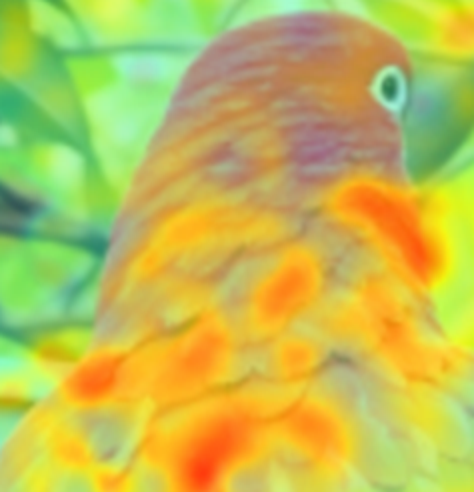} &
    \includegraphics[width=\imgw]{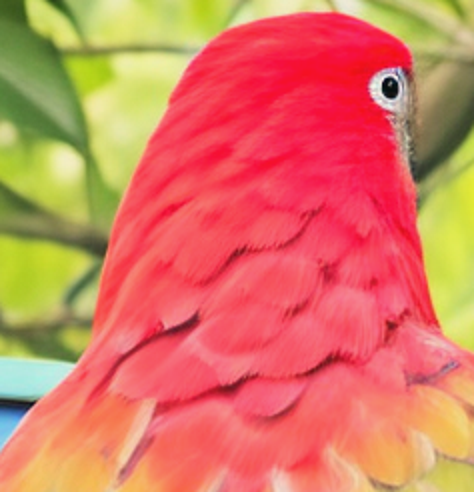} &
    \includegraphics[width=\imgw]{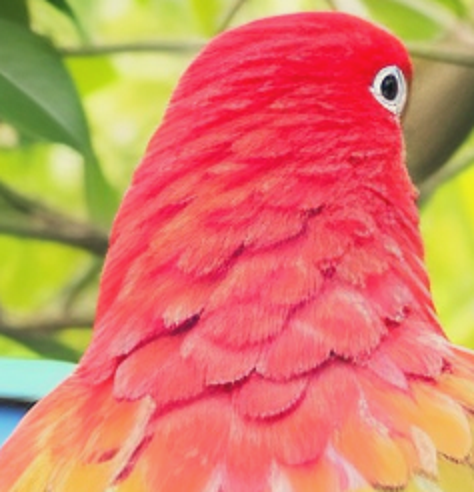} \\[-3pt]
    \small (a) LQ patch &
    \small (b)  &
    \small (c) &
    \small (d) OSEDiff~\cite{OSEDiff} &
    \small (e) Ours \\
  \end{tabular}}
  \vspace{-2mm}
  \caption{
    Visualization comparison of DAAMs~\cite{DAAM} for the query word \textit{“bird”}. 
    (b) and (c) are DAAMs for OSEDiff~\cite{OSEDiff} and our method.
    Compared to~\cite{OSEDiff}, our method achieves more precise text–image interaction in the semantic region corresponding to the bird, resulting in more realistic feather generation.
  }
  \label{fig:ablation_tmgs}
  \vspace{-6mm}
\end{figure*}

\begin{figure}[t]
    \scriptsize
    \centering
    \setlength{\tabcolsep}{2pt} 
    \begin{tabular}{cc}
        \includegraphics[width=0.90\columnwidth]{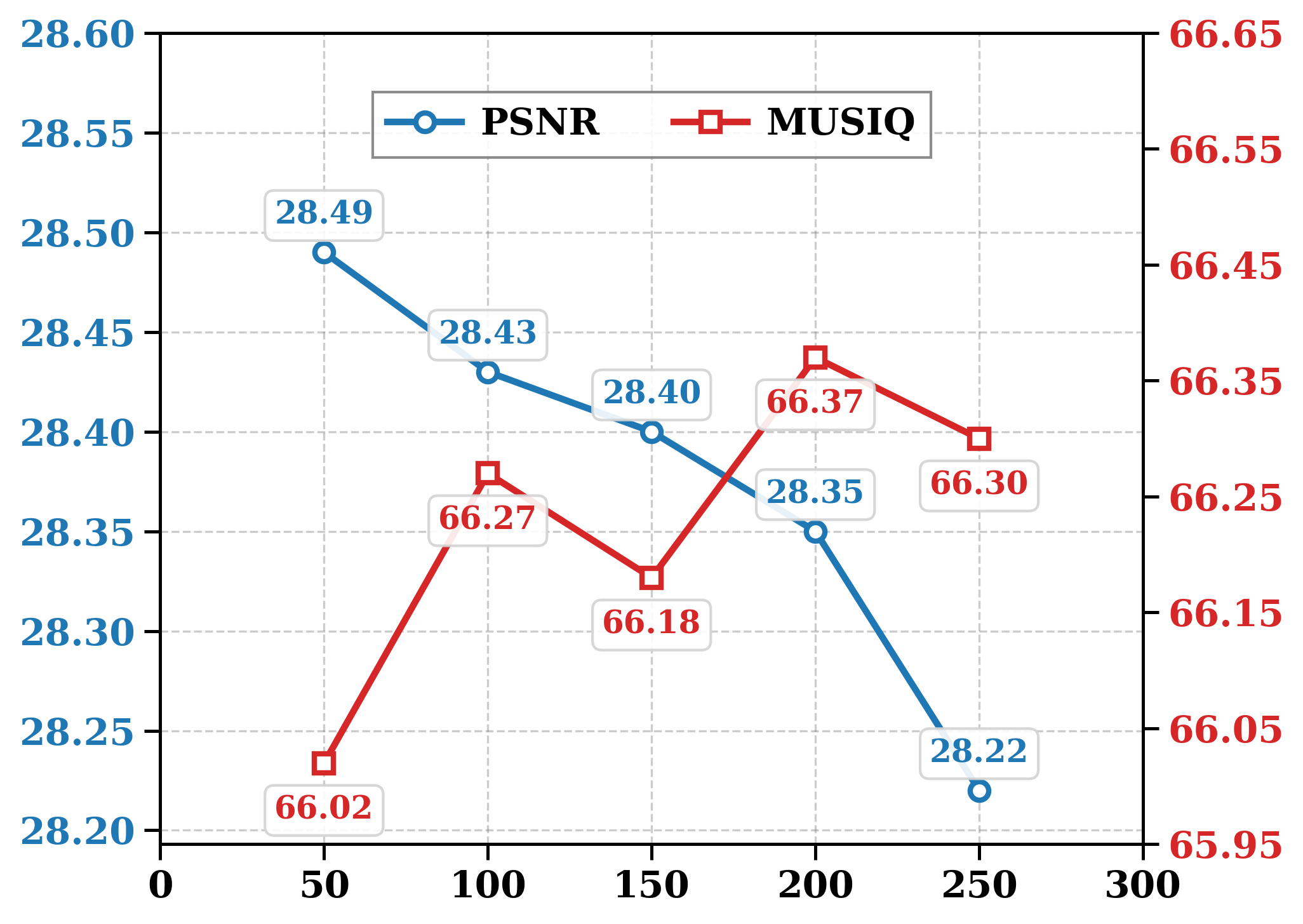} \\
    \end{tabular}
    \vspace{-2mm}
    \caption{
    PSNR and MUSIQ metrics variation with different timestep on the DrealSR~\cite{DrealSR} benchmark. 
    }
    \label{fig:ablation_intensity}
    \vspace{-4mm}
\end{figure}

We additionally investigate the impact of adaptive noise $\bm{\epsilon}_a$ intensity on reconstruction by adjusting the timestep $t_s$ during the forward process.
The trend in \Cref{fig:ablation_intensity} demonstrates a clear trade-off between fidelity and quality: higher $t_s$ values (i.e., greater noise intensity) enhance the release of generative priors, as illustrated in the increasing MUSIQ values, but concurrently degrade fidelity, as evidenced by the decrease in PSNR.
This property is characterized by a transition from fidelity-preserving reconstruction at lower noise levels to perceptually-oriented generation at higher levels.
To balance these competing objectives, we set $t_s = 100$ as the default setting for subsequent experiments.


\noindent
\textbf{Effect of LQ-guided feature modulation.} 
To validate the effectiveness of our LQFM, we compare \emph{Base$_{\text{w/ RGPA}}$} with a variant that is augmented with the LQFM module (\emph{Base$_\text{w/ RGPA\&LQFM}$} for short).
As shown in Table~\ref{tab:ablation_fmm}, using LQFM yields substantial improvements in both fidelity and perceptual metrics.
We further compare \emph{Base$_\text{w/ RGPA\&LQFM}$} with two variants that respectively replace the modulation in \eqref{eq: SFT_modulation} with the element-wise addition $\bm{f}^m+\text{MLP}(\bm{\widetilde{x}}_L)$ (\emph{Base$_{\text{w/ RGPA\&}\bm{f}^m+\text{MLP}(\bm{\widetilde{x}}_L)}$} for short) or SFT($\bm{z}_L \mid\bm{\widetilde{x}}_L$) (\emph{Base$_{\text{w/ RGPA\&}\text{SFT}(\bm{z}_L \mid\bm{\widetilde{x}}_L)}$} for short).
The comparison results in Table~\ref{tab:ablation_fmm} show that using element-wise addition performs suboptimally, even degrading no-reference metrics compared to \emph{Base$_{\text{w/ RGPA}}$}. 
We attribute this to the domain gap between the LQ input and U-Net features, which introduces instability into the denoising process and can lead to its eventual collapse of the denoising paradigm.
Moreover, modulating the latent feature $\bm{z}_L$ causes a more pronounced drop in no-reference metrics than modulating the intermediate feature $\bm{f}^m$, as it disrupts the Gaussian diagonal distribution of $\bm{z}_L$ and consequently limits the effective utilization of generative priors during the reverse process.
In contrast, our proposed LQFM effectively enhances restoration fidelity while maintaining better generative performance.

\begin{table}[!t]
  \centering
  \caption{Comparison of different modulation strategies on the DrealSR~\cite{DrealSR} benchmark.}
  \label{tab:ablation_fmm}
  \vspace{-2mm}
  \resizebox{1.0\linewidth}{!}{
  \begin{tabular}{@{}l c c c c c@{}}
    \toprule
                                & PSNR $\uparrow$ & FID $\downarrow$ & MUSIQ $\uparrow$ & CLIPIQA+ $\uparrow$ \\
    \midrule
    $\text{Base}_\text{w/ RGPA}$ & 28.24 & 133.02 & 66.26 & 0.5109
            \\
   $\text{Base}_{\text{w/ RGPA\&}\bm{f}^m+\text{MLP}(\bm{\widetilde{x}}_L)}$  & 28.40 & 132.76              & 65.75            & 0.4956            \\
    $\text{Base}_{\text{w/ RGPA\&}\text{SFT}(\bm{z}_L \mid\bm{\widetilde{x}}_L) }$  & 28.24                      & 128.64              & 65.82           & 0.5065           \\
    $\text{Base}_\text{w/ RGPA\&LQFM}$  & \textbf{28.43} & \textbf{128.34} & \textbf{66.27} & \textbf{0.5213}    \\
    \bottomrule
  \end{tabular}}
  \vspace{-7mm}
\end{table}



\begin{table}[!t]
  \centering
  \caption{Comparison of different semantic enhancement strategies on the DrealSR~\cite{DrealSR} benchmark.}
  \label{tab:ablation_TMG}
  \vspace{-2mm}
  \resizebox{1.0\linewidth}{!}{
  \begin{tabular}{@{}l c c c c c@{}}
    \toprule
                                & NIQE $\downarrow$ & MUSIQ $\uparrow$ & MANIQA $\uparrow$ & CLIPIQA+ $\uparrow$ \\
    \midrule
    $\text{Base}_\text{w/ RGPA\&LQFM}$ & 6.00 & 66.27 &\textbf{0.6317} & 0.5109 \\
    %
    $\text{Ours}_{\text{w/o VSD loss}}$  & 6.07  & 65.56  & 0.6257  & 0.5009  \\
    $\text{Ours}$  & \textbf{5.97} & \textbf{67.05} & 0.6278 & \textbf{0.5589}    \\
    \bottomrule
  \end{tabular}}
  \vspace{-7mm}
\end{table}

\noindent
\textbf{Effect of text-matching guidance.} 
To demonstrate the effectiveness of the TMG, we compare our full model with \emph{Base$_\text{w/ RGPA\&LQFM}$}. 
As shown in Table~\ref{tab:ablation_total}, our approach with TMG yields further improvements across all non-reference metrics, underscoring its critical role in enhancing the generation capability of the model. 
\Cref{fig:ablation_tmgs} visualizes diffusion attentive attribution maps (DAAMs) of our method and~\cite{OSEDiff}, where the method~\cite{OSEDiff} suffers from text misalignment even with the regularization of pre-trained models. 
In contrast, our proposed TMG effectively alleviates the misalignment issue, leading to more accurate and consistent text–image interactions.
%

To further investigate the influence of semantic knowledge distillation in the second training stage, we compare our CODSR with a baseline that replaces the VSD loss in \eqref{eq:loss_stage2} with the GAN loss (\emph{Ours$_{\text{w/o VSD loss}}$} for short).
As shown in Table~\ref{tab:ablation_TMG}, 
using GAN Loss fails to fully unleash the semantic generation capability due to the lack of effective semantic guidance in the second-stage training, whereas the VSD Loss enables semantic knowledge distillation from the pretrained model, thereby enhancing the generative ability.
%
%

\section{Conclusion}
We present CODSR, a diffusion-based one-step SR framework in order to achieve a favorable balance between structural fidelity and perceptual quality.
We introduce spatially adaptive noise to the LQ latent, enabling region-discriminative activation of generative priors.
We further develop a lightweight LQ-guided feature modulation module to preserve information from the LQ image and mitigate the LQ information loss caused by compression encoding, thus enhancing fidelity without compromising generative capability.
In addition, we propose a text-matching guidance strategy to provide explicit spatial guidance for text-interactive regions, achieving precise semantic grounding.
Extensive experiments on benchmarks demonstrate that CODSR outperforms state-of-the-art methods in terms of structural fidelity and visual quality.

{
    \small
    \bibliographystyle{ieeenat_fullname}
    \bibliography{main}
}


\end{document}